# TRANSFER LEARNING FOR DIAGNOSIS OF CONGENITAL ABNORMALITIES OF THE KIDNEY AND URINARY TRACT IN CHILDREN BASED ON ULTRASOUND IMAGING DATA

*Qiang Zheng[1,2,3], Gregory Tasian[4], Yong Fan[1]*

[1]Department of Radiology, School of Medicine, University of Pennsylvania, Philadelphia, PA, 19104, USA
[2]School of Computer and Control Engineering, Yantai University, Yantai, 264005, China
[3]National Laboratory of Pattern Recognition, Institute of Automation, Chinese Academy of Sciences, Beijing, 100190, China
[4]The Children's Hospital of Philadelphia, Philadelphia, PA, 19104, USA

## ABSTRACT

Classification of ultrasound (US) kidney images for diagnosis of congenital abnormalities of the kidney and urinary tract (CAKUT) in children is a challenging task. It is desirable to improve existing pattern classification models that are built upon conventional image features. In this study, we propose a transfer learning-based method to extract imaging features from US kidney images in order to improve the CAKUT diagnosis in children. Particularly, a pre-trained deep learning model (imagenet-caffe-alex) is adopted for transfer learning-based feature extraction from 3-channel feature maps computed from US images, including original images, gradient features, and distanced transform features. Support vector machine classifiers are then built upon different sets of features, including the transfer learning features, conventional imaging features, and their combination. Experimental results have demonstrated that the combination of transfer learning features and conventional imaging features yielded the best classification performance for distinguishing CAKUT patients from normal controls based on their US kidney images.

## 1. INTRODUCTION

Children with congenital abnormalities of the kidney and urinary tract (CAKUT) face many challenges and early detection of CAKUT can help prevent the progression of kidney disease to kidney failure [1]. Ultrasound (US) imaging plays a key role in CAKUT diagnosis because anatomic measures derived from US imaging data, such as renal elasticity, renal parenchymal area, maximum renal length, and cortical thickness, are correlated with kidney function [2, 3]. More recently, texture analysis of imaging data has demonstrated improved performance for predicting change in kidney function [4].

Pattern recognition methods have been used to aid kidney disease diagnosis based on US imaging data. Particularly, a decision support system has been developed to classify US images of normal controls and renal disease patients based on second order grey level co-occurrence matrix statistical features [5]; neural networks in conjunction with principal component analysis have been used to classify US kidney images [6], and support vector machine (SVM) classifiers have been built upon texture features extracted from regions of interest of US images to classify kidney images [7, 8]. These studies have demonstrated that pattern classifiers built upon imaging features could obtain promising performance for classifying US imaging data.

The success of deep learning techniques in recent years have witnessed promising performance in learning imaging features for a variety of pattern recognition tasks [9-11]. In these studies, convolutional neural networks (CNNs) are widely adopted to learn informative imaging features by optimizing pattern recognition cost functions. Therefore, it is expected that incorporating the deep learning techniques into US imaging data analysis would improve US image classification and subsequently improve diagnosis of CAKUT in children based on US imaging data.

Since a large dataset is typically needed to build a generalizable deep learning based classification model [12], for applications with small datasets the deep learning tools are often adopted in a transform learning setting, i.e., applying deep learning models trained based on a large dataset for one problem to a different but related problem with a relatively small training dataset [13]. The transfer learning strategy has achieved promising performance in pattern recognition studies of medical image data [14-19]. Furthermore, it has been demonstrated that classifiers built upon combination of transfer learning features and hand-crafted features typically achieve better pattern recognition performance than those built upon transfer learning features and hand-crafted features alone [17-19].

Building upon the successful deep learning and transfer learning techniques, we develop a pattern recognition method for distinguishing children with CAKUT from healthy kids based on their US kidney imaging data. Particularly, we adopt a pre-trained model of CNNs, namely imagenet-caffe-alex [20], for extracting imaging features from 2D US kidney images. Since the US kidney images have only one channel of intensity values while the imagenet-caffe-alex model requires color images with 3 channels, two more feature maps are computed from each original US image, including a gradient feature map and a distanced transformation feature map, and they are integrated with the original US image to be used as a 3-channel input to the imagenet-caffe-alex model. Moreover, hand-crafted texture features are also extracted from each US image. Finally, SVM classifiers are built upon the transfer learning features and hand-crafted texture features to distinguish US kidney images of children with CAKUT from those of healthy kids. We have validated our method based on an US imaging dataset obtained from 50 children with CAKUT and 50 healthy kids at the Children's Hospital of Philadelphia (CHOP). Experiment results have demonstrated that our method could achieve promising pattern classification performance for aiding diagnosis of CAKUT.

## 2. METHODS

Our method consists of 3 components, including kidney segmentation, feature extraction, and SVM based classification, as illustrated in Fig. 1.

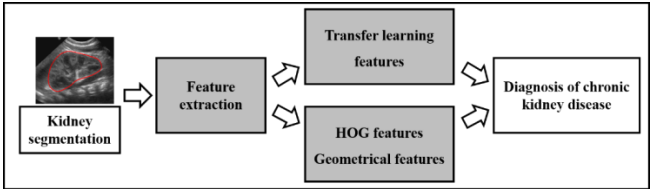

**Fig.1** Flowchart for CAKUT diagnosis based on US images.

### 2.1 Kidney Segmentation and image normalization

We adopted a graph-cuts method to segment kidneys in US images [21]. Particularly, the graph-cuts method segments kidneys in US images based on both image intensity information and texture features extracted using Gabor filters. This method has achieved promising segmentation performance for segmenting bilateral kidneys in US images. The average Dice index of the segmentation results compared with manual segmentation results was 0.96.

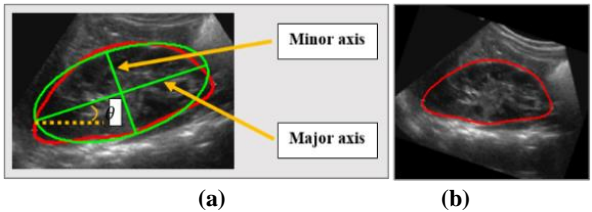

**Fig. 2** Ellipse estimation (a) and image rotation (b).

To make US kidney images of different subjects directly comparable, we normalized US kidney images of different subjects. First, the orientation of kidneys in US images is estimated based on ellipse fitting, including the ellipse's major axis, minor axis, and the orientation $\theta$ between the major axis and X-axis, as illustrated in Fig. 2. Second, based on the estimated ellipse information, each US kidney image is reoriented along the major axis and resized to have the same size as the images used in the deep learning model to be described. Particularly, the reoriented image is centered to the ellipse's center and rescaled according to the ellipse's major axis and minor axis to have the same size. Finally, image intensity values outside of the kidney are set to zero.

### 2.2 Feature Extraction

Two different methods are used to extract image features from the normalized kidney images, including the transfer learning based on a deep CNN model and conventional texture features.

*2.2.1 Feature Extraction by Transfer Learning*

A pre-trained CNN model (imagenet-caffe-alex) is adopted from MatConvNet to extract features from the kidney images [20]. The imagenet-caffe-alex model was trained on 1.2 million 3-channle images of the ImageNet LSVRC-2010 for classifying images into 1000 different classes. Since the kidney US images only have one channel of intensity values, we propose to adopt original kidney image, gradient image, and distanced transform image to build a 3-channel image, as illustrated in Fig. 3.

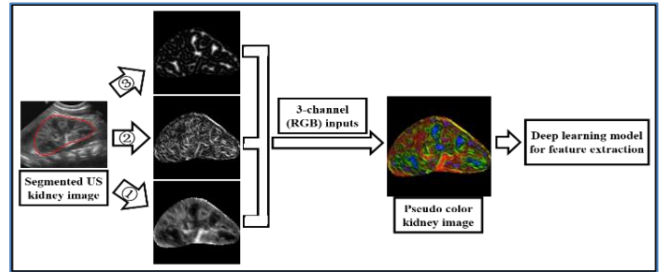

**Fig. 3** Feature extraction by transfer learning. ① original US kidney image; ② gradient image; ③ distanced transform image.

In particular, 3 feature maps are computed from each normalized US kidney image, including original kidney US image $f_I(x,y)$, gradient feature map $f_G(x,y)$, and distanced transform feature map $f_D(x,y)$. First, the original US kidney image intensity values are normalized to [0, 255]. Second, a gradient feature map $f_G(x,y)$ is computed as:

$$f_G(x,y) = \frac{g(x,y)}{f_I(x,y)} = \frac{\sqrt{g_x^2(x,y)+g_y^2(x,y)}}{f_I(x,y)}, \quad (1)$$

where $g_x(x,y) = (f_I(x+1,y) - f_I(x-1,y))/2$, $g_y(x,y) = (f_I(x,y+1) - f_I(x,y-1))/2$, $x$ and $y$ are coordination of pixels. Third, a distance transform feature map is computed using the VLfeat toolbox [22] by

$$f_D(x,y) = \min_{x',y'}(f_I(x',y') + ((x-x')^2 + (y-y')^2)). \quad (2)$$

Particularly, the distance transform map characterizes the distance of each pixel to its nearest element in an edge map obtained by applying Canny edge detector to the original image [23]. Both the gradient and distance transformation feature maps are normalized to [0, 255].

Finally, each kidney image's original kidney US image $f_I(x,y)$, gradient feature map $f_G(x,y)$, and distanced transform feature map $f_D(x,y)$ are used as R-channel, G-channel, and B-channel respectively to form a 3-channel image so that the imagenet-caffe-alex model could be adopted to extract features. Two randomly selected kidney US images, their RGB channels, and their pseudo color images are shown in Fig. 4.

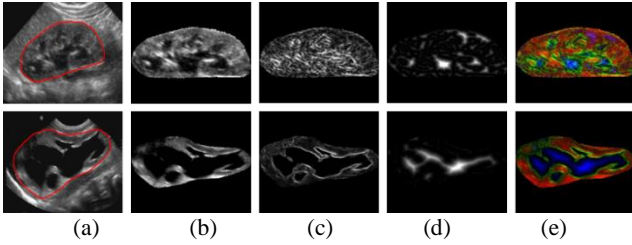

**Fig.4** US kidney images and feature maps. (a) US images with kidney contours in red; (b) original image feature $f_I(x,y)$; (c) gradient feature $f_G(x,y)$; (d) distanced transform feature $f_D(x,y)$, and (e) pseudo color images.

### 2.2.2 Conventional Image Features

Conventional image features are also extracted from the US kidney images, including histogram of oriented gradients (HOG) features and geometrical features of the kidneys.

(1) HOG Features

The HOG decomposes an image into small squared cells, computes the histogram of oriented gradients in each cell, normalizes the result using a block-wise pattern, and returns a descriptor for each cell. Same as the distanced transform, we utilize the VLFeat toolbox [22] to compute the HOG features with default parameters except that the cell size is set to $N_0/10$, where $N_0$ is the size of the input images to the deep learning model.

(2) Geometrical Features

The geometrical features include shape-related measures and block-related measures of the kidneys. Particularly, the shape-related features are defined as $V_{shape} = [L_1, L_2, L_1/L_2, L_1 \times L_2, L_1 + L_2, L_1^2 + L_2^2, L_1 - L_2, L_1^2 - L_2^2]$, where $L_1$ and $L_2$ are lengths of the major and minor axes of the kidney as defined by the ellipse estimation of the kidney. The block-related features are ratios of areas of black holes inside kidney to the whole kidney region. Because no suitable threshold is available for segmenting holes from all kidney images, we uniformly set 10 thresholds of $[3:3:30]$ to segment black holes, and the resulting ratios constitute the hold-related features $V_{block} = [ratio_1, \cdots, ratio_{10}]$. Finally, from each kidney image, we obtain a set of geometric features $V_{geometric} = [V_{shape}, V_{block}]$.

### 2.3 Diagnosis of CAKUT

The diagnosis of CAKUT based on US kidney images is modeled as a pattern classification problem based on US image features. Particularly, a L2-regularized L1-loss SVM is utilized to build classifiers based on the extracted image features by optimizing

$$\min_{w} \tfrac{1}{2}\vec{w}^T\vec{w} + C \sum_{i=1}^{l}\left(\max(0, 1 - y_i\vec{w}^T\vec{f_i})\right) \quad (5)$$

where $\vec{f_i}$ is the feature vector of the $i^{th}$ US kidney image, $\vec{w}$ is the weighting vector to be learned from training data.

The L2-regularized L1-loss SVM optimization problem can be solved using a dual coordinate descent method. Particularly, a publicly available software package LIBLINEAR with its default parameters is utilized to build the SVM classifiers [24]. Once we get the weighting vector $\vec{w}$, the category of the $i^{th}$ kidney image can be estimated by

$$L_i = sgn(\vec{w}^T\vec{f_i}). \quad (6)$$

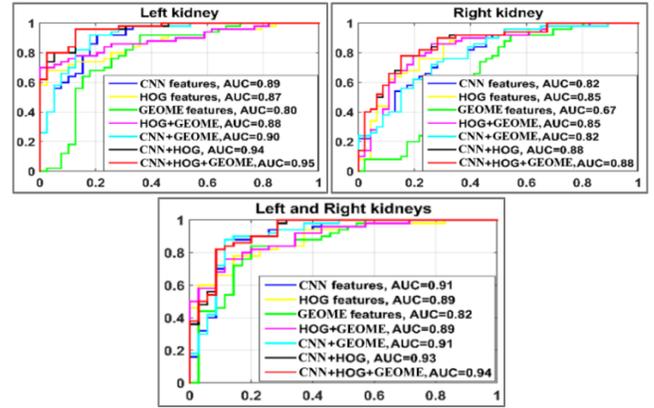

**Fig.5** ROC curves of different classifiers, estimated based one run of 10-fold cross-validation. GEOME: geometrical features.

## 3. EXPERIMENTAL RESULTS

We have validated our method based on kidney images collected at the Children's Hospital of Philadelphia. The dataset contains kidney images obtained from 50 normal subjects and 50 patients. Particularly, all the 50 normal subjects had both left and right kidney images, 35 patients had abnormal kidneys in both sides, 39 patients had abnormal kidneys in the left side, and 46 patients had abnormal kidneys in the right side. We built SVM classifiers based on all the available data for the left, right, and bilateral kidney images separately, and 10-fold cross-validation was adopted to evaluate the classification performance. The 10-fold cross-validation was repeated for 100 times to estimate the classification performance.

In order to investigate if the transfer learning features and the conventional image features provide complementary information for distinguishing abnormal and normal kidney images, we compared classifiers built upon different sets of features, including the transfer learning features alone, the conventional features alone, and their combination.

As shown in Fig.5, the classifiers built upon the combination of the transfer learning features and the conventional features had the largest area under the receiver operating characteristic (ROC) curve (AUC), indicating that integrating the transfer learning features and the conventional imaging features could improve the classification of US kidney images. Table 1 summarizes mean classification accuracy and AUC values of 100 runs of the 10-fold cross-validation experiments. These results further demonstrated the advantages of combination (CNN+HOG+Geometrical features) over others.

Table 1. 10-fold cross-validation (mean±standard*e-2).

|  | CNN | HOG | GEOM | HOG + GEOME | CNN + GEOME | CNN + HOG | CNN +HOG +GEOME |
|---|---|---|---|---|---|---|---|
| classification accuracy | | | | | | | |
| Left | 0.85±1.8 | 0.77±2.3 | 0.78±1.5 | 0.79±2.7 | 0.87±1.5 | 0.84±1.4 | **0.84±1.6** |
| Right | 0.75±2.0 | 0.74±2.5 | 0.67±1.7 | 0.74±2.4 | 0.74±2.1 | 0.81±1.8 | **0.81±1.8** |
| Both | 0.85±1.2 | 0.78±2.4 | 0.76±1.8 | 0.79±2.3 | 0.85±1.4 | 0.86±2.1 | **0.87±2.1** |
| area under the ROC curve | | | | | | | |
| Left | 0.88±0.8 | 0.84±1.5 | 0.80±0.9 | 0.86±1.5 | 0.88±0.9 | 0.92±0.8 | **0.92±0.7** |
| Right | 0.83±1.3 | 0.83±1.6 | 0.67±1.7 | 0.83±1.6 | 0.83±1.2 | 0.88±1.0 | **0.88±1.1** |
| Both | 0.89±0.8 | 0.87±1.6 | 0.82±1.3 | 0.87±1.5 | 0.89±0.8 | 0.92±0.7 | **0.92±0.7** |

GEOME: Geometrical features.

## 4. CONCLUSIONS

In this paper, we proposed a transfer learning-based method for CAKUT diagnosis based on US kidney images. The classification experiments for distinguishing CAKUT patients from normal controls based on their kidney US images have demonstrated that integrating the transfer learning features and conventional image features could improve the classification of US kidney images.

## ACKNOWLEDGEMENTS

This work was supported in part by the National Key Basic Research and Development Program of China [2015CB856404]; the National Natural Science Foundation of China [61473296]; the Promotive Research Fund for Excellent Young and Middle-Aged Scientists of Shandong Province [BS2014DX012]; China Postdoctoral Science Foundation [2015M581203]; the International Postdoctoral Exchange Fellowship Program [20160032]; and National Institutes of Health grants [EB022573, MH107703, DK114786, DA039215, and DA039002].